\pdfoutput=1
\documentclass[final]{cvpr}

\usepackage{times}
\usepackage{epsfig}
\usepackage{graphicx}
\usepackage{amsmath}
\usepackage{amssymb}

\usepackage{amsmath}
\usepackage{booktabs}
\usepackage{multirow}
\usepackage{amssymb}
\usepackage{algorithm, algorithmic}

\usepackage[pagebackref=true,breaklinks=true,colorlinks,bookmarks=false]{hyperref}

\def\cvprPaperID{3365} 
\def\confYear{CVPR 2021}

\begin{document}

\title{AACP: Model Compression by Accurate and Automatic Channel Pruning}

\author{Lanbo Lin, Yujiu Yang, Zhenhua Guo\\
Tsinghua University\\
{\tt\small llb19@mails.tsinghua.edu.cn, yang.yujiu@sz.tsinghua.edu.cn, zhenhua.guo@sz.tsinghua.edu.cn}
}

\maketitle

\begin{abstract}
   Channel pruning is formulated as a neural architecture search (NAS) problem recently. However, existing NAS-based methods are challenged by huge computational cost
   and inflexibility of applications. How to deal with multiple sparsity constraints simultaneously and speed up NAS-based 
   channel pruning are still open challenges. In this paper, we propose a novel Accurate and Automatic Channel Pruning 
   (AACP) method to address these problems. Firstly, AACP represents the structure of a model as a structure vector and introduces a
   pruning step vector to control the compressing granularity of each layer. Secondly, AACP utilizes Pruned Structure Accuracy Estimator (PSAE) to
   speed up the performance estimation process. Thirdly, AACP proposes Improved Differential Evolution (IDE) 
   algorithm to search the optimal structure vector effectively. Because of IDE, AACP can deal with FLOPs constraint and model size constraint simultaneously and efficiently.
   Our method can be easily applied to various tasks and achieve state of the art performance. On CIFAR10, our method reduces $65\%$ FLOPs 
   of ResNet110 with an improvement of $0.26\%$ top-1 accuracy. On ImageNet, we 
   reduce $42\%$ FLOPs of ResNet50 with a small loss of $0.18\%$ top-1 accuracy and reduce $30\%$ FLOPs of MobileNetV2 with a small loss of $0.7\%$ top-1 accuracy. 
   The source code will be released after publication.
\end{abstract}

\begin{figure}[t]
   \centering
   \includegraphics[width=1\columnwidth]{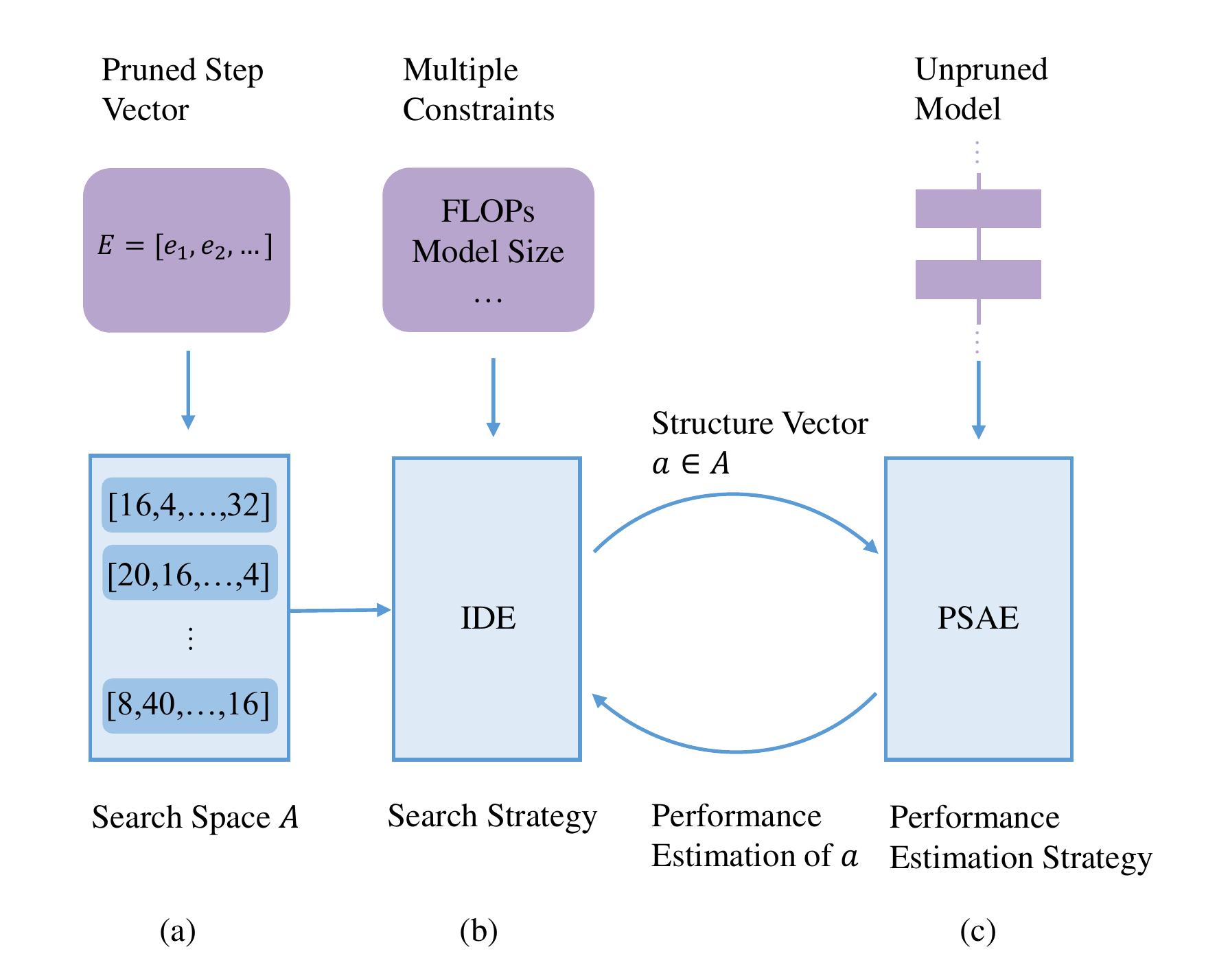} 
   \caption{
       Framework of AACP. (a) Search space. We represent the architecture of a CNN as a vector of its channel numbers. 
       All possible combinations of channel numbers form the search space. A pruned step vector is adopted to compress the search space.
       (b) Search strategy. AACP proposes Improved Differential Evolution (IDE) algorithm to search an optimal pruned structure. IDE can find the optimal structure under 
       multiple constraints. (c) Performance Estimation Strategy. Given a structure vector $a$, AACP adopts Pruned Structure Accuracy Evaluator (PSAE) to estimate its performance.
       A pre-trained unpruned model is utilized to share weights with pruned models.
   }
   \label{fig1}
\end{figure}

\section{Introduction}
Convolutional Neural Networks (CNNs) have made great progress in computer vision tasks
\cite{he2016deep,long2015fully,ren2015faster,chen2017deeplab}. However, a deep CNN
usually has high demands in computing power and memory consumption. For cell phones or other mobile devices, 
those CNNs are unaffordable. Hence, model compression \cite{lecun1990optimal,han2015deep,hinton2015distilling} is necessary
for applications of CNNs. Neural network pruning \cite{han2015deep,li2016pruning} is an effective way of model compression, which aims to reduce
FLOPs and redundant parameters of a CNN while keeping its high performance.

Recently, channel pruning has become a hotspot in neural network pruning because of its effectiveness in accelerating inference time of a CNN. 
A recent work \cite{liu2018rethinking} proposes that fine-tuning 
a pruned model only gives comparable or worse performance than training that model with randomly initialized weights. This finding 
implies the pruned architecture is more important than inheriting “important”  weights from an unpruned model. Inspired by this 
discovery, some works \cite{guo2020dmcp,yu2019autoslim,lin2020channel,liu2019metapruning} focus on searching the optimal 
architecture from an unpruned model.

AMC \cite{he2018amc} utilizes reinforcement learning to search the pruning ratio of each layer. MetaPruning \cite{liu2019metapruning}
trains a large PruningNet as an one-shot model to generate weights for pruned networks. Then evolution algorithm is adopted to search for 
efficient architectures. ABCPruner \cite{lin2020channel} proposes to use artificial bee colony algorithm as the search strategy. 
Instead of training an one-shot model to estimate the performance of an architecture, ABCPruner chooses to fine-tune every architecture,
which is computationally expensive. AutoSlim \cite{yu2019autoslim} trains a slimmable network to approximate the performance of different architectures 
and slims the network greedly. These methods are computationally expensive either in the process of training a large one-shot model or in 
the process of searching optimal architectures. DMCP \cite{guo2020dmcp} models channel pruning as a differentiable Markov process,
which eliminates the effort of searching a large amount of architectures. However, differentiable modeling makes it harder to optimize
the pruning problem when there are multiple constraints simultaneously, such as constraints of FLOPs, model size and inference time.

In this paper, we propose a novel Accurate and Automatic Channel Pruning (AACP) method to alleviate these problems (as shown in Fig.\ref{fig1}). Firstly, 
we represent the structure of a model as a vector, whose elements stand for the channel numbers. 
We propose a pruning step vector $E$ to compress the searching space and accurately control the pruning granularity of each layer. Secondly, 
we improve Differential Evolution algorithm to search for
optimal structures fastly. Our Improved Differential Evolution (IDE) algorithm can help to avoid getting stuck in
locally optimal solutions and deal with multiple constraints simultaneously. Thirdly, we propose a novel performance estimation strategy named 
Pruned Structure Accuracy Evaluator (PSAE) to speed up the process of performance estimation, which is faster than utilizing an one-shot model.

Our main contributions can be summarized as:
\begin{itemize}
\item We propose a novel channel pruning method AACP to prune CNNs in an end-to-end manner, which achieves state-of-the-art 
performance in many pruning tasks.
\item Our method largely speeds up the process of performance estimation because of the proposed Pruned Structure Accuracy Evaluator (PSAE). 
PSAE utilizes $l_1$-norm criterion to select weights, eliminating the efforts of training an one-shot model or fine-tuning a given architecture. 
So our method is simpler than other NAS-based methods.
\item Our method can achieve accurate pruning under multiple constraints simultaneously because of the proposed Improved Differential Evolution (IDE) 
algorithm. We can flexibly control the pruning rates of FLOPs, model size and inference time at the same time.
\end{itemize}    

\section{Related Works}
Neural network pruning can be summarized into weight pruning and channel pruning. Weight pruning methods \cite{lecun1990optimal,han2015deep}
focus on pruning fine-grained weights, which cannot be accelerated directly without other software supports. In contrast, channel 
pruning methods remove the whole filters, leading to structure sparsity. There are three main categories of channel pruning:

\textbf{Importance metric.} Some works utilize certain metrics to evaluate the importance of filters. \cite{li2016pruning,he2018soft} propose to use $l_1$-norm to sort filters. 
\cite{hu2016network} prunes those filters 
with lower Average Percentage of Zeros(APoZ) after the ReLU mapping. \cite{molchanov2016pruning} uses Taylor expansion to approximate 
the change in loss function induced by pruning some filters. The filters that cause less change in loss function are thought to be less important 
than other filters. \cite{luo2017thinet} prunes filters based on statistics information computed from its next layer. \cite{he2019filter} 
calculates the geometric median of the filters within the same layer and 
removes those ones that are closest to the geometric median. \cite{lin2020hrank} proposes to prune filters with low-rank feature maps. 
\cite{he2020learning} utilizes different metrics to evaluate the importance of filter in different layers. Most pruning methods based on 
importance metrics focus on selecting "important" filters instead of finding optimal channel configurations. They have the advantage of low 
time complexity, but also have limits in performance and compressing rate of pruning results. 

\textbf{Sparsity Regularization.} Another idea is to impose sparsity regularization on filters or channels by changing the loss function. 
\cite{liu2017learning} proposes to impose sparsity constraints on 
the scaling factors of batch normalization layers. After training is finished, those channels with small scaling factors will 
be pruned. \cite{huang2018data} introduces scaling factors to scale the outputs of specific structures, including filters, groups and residual 
blocks. Some works impose sparsity constraints directly on weights instead of scaling factors \cite{wen2016learning,alvarez2016learning}. 
\cite{ding2019centripetal} proposes 
centripetal SGD to train several filters identical and keeps only one of them at the end. \cite{luo2020autopruner} adds AutoPruner block to 
each convolutional layer and forces the activation of some channels to be zero. However, since the loss functions of sparsity constraints are 
typically not differentiable, it takes a lot of efforts to optimize a network under multiple sparsity constraints.
In contrast, our AACP doesn't add sparsity regularization to the loss function, so we can handle multiple constraints simultaneously and easily.

\textbf{Neural Architecture Search(NAS).} Some works focus on searching the best structure, i.e. channel number in each layer, instead 
of evaluating the importance of filters. \cite{wang2019pruning} points out that a 
pre-trained over-parameterized model is not necessary for searching the efficient architectures. \cite{he2018amc} utilizes reinforcement 
learning to search the channel numbers layer by layer. \cite{yu2019autoslim,liu2019metapruning} firstly train an one-shot model and then 
search the efficient architectures within the one-shot model. \cite{lin2020channel} finds an optimal pruned structure based on artificial bee 
colony algorithm. \cite{guo2020dmcp} makes the channel pruning differentiable by modeling it as a Markov process. These methods have high
computational cost either in the process of training a large one-shot model \cite{yu2019autoslim,liu2019metapruning,he2018amc} 
or in the process of searching an optimal architecture \cite{lin2020channel}. Our method dosen't need to train an one-shot model or 
fine-tune pruned models in searching stage, which largely speeds up the process of finding the optimal structures.

\section{Methodology}
\subsection{Definition of channel pruning}

Given an unpruned model $\mathcal{M}$ with $L$ convolutional layers, we can represent the architecture of $\mathcal{M}$ as a vector 
$C=[c_1, c_2,..., c_L]$, where $c_i$ refers to the output channel number of the $i$-th convolutional layer. We 
denote a pruned model as $\mathcal{M^{'}}$ and the architecture of $\mathcal{M^{'}}$ as $C^{'}=[c_{1}^{'}, c_{2}^{'},
..., c_{L}^{'}]$. Note that $0<c_{i}^{'} \leq c_{i}, c_{i}^{'} \in \mathbb{Z}$.

The goal of channel pruning is to find an optimal structure vector $C^{'*}=[c_{1}^{'*}, c_{2}^{'*},
..., c_{L}^{'*}]$ under certain sparsity constraints. 
We introduce $r_f \in [0,1]$ as the pruning rate of FLOPs and $r_p \in [0,1]$ 
as the pruning rate of parameters. Other kinds of constraints, such as inference time, can be dealt with in the same way. 
Given a pruned model $\mathcal{M^{'}}$ with structure $C^{'}$ and an unpruned model $\mathcal{M}$ with structure $C$, 
we have:
\begin{equation}
\begin{split}
    r_f(C^{'}) = 1 - \frac{FLOPs(C^{'})}{FLOPs(C)} \\
    r_p(C^{'}) = 1 - \frac{Params(C^{'})}{Params(C)}\label{eq1}
\end{split}
\end{equation}

\noindent where $FLOPs(*)$ is the calculation function of FLOPs and $Params(*)$ is the calculation function of parameters. We have $0 \leq r_f(*) \leq 1$ 
and $0 \leq r_p(*) \leq 1$. Therefore, our channel pruning problem can be formulated as:
\begin{align}
    C^{'*} = &\mathop{max} \limits_{C^{'}}(acc(C^{'};C)) \notag\\
    s.t. \quad &r_f(C^{'}) \geq R_{f} \notag\\
    &r_p(C^{'}) \geq R_{p} \label{eq2}
\end{align}

\noindent where $C^{'*}$ is the optimal structure vector, $R_{f}$ is the target of FLOP pruning rate, $R_{p}$ is the 
target of parameter pruning rate and $acc(*;C)$ is the validation accuracy of a model obtained by pruning $C$.

\subsection{Compressing search space}

Searching for the optimal structure vector $C^{'*}$ is a nonlinear optimization problem. Brute-force 
method is not available due to its huge amount of computation. For example, the number of all possible structure vectors of a $L$-layer 
CNN is $\prod_{i=1}^{L}c_i$, which is of exponential growth rate. In order to alleviate this problem, we need to compress 
the search space.
The search space of a CNN is composed of all possible structure vectors, which can be denoted as $\mathcal{S} \in \mathbb{R}^L$. We have:
\begin{equation}
    \mathcal{S} = \mathcal{D}_1 \times \mathcal{D}_1 \times ... \times \mathcal{D}_L \label{eq3}
\end{equation}

\noindent where $\mathcal{D}_i=\{1,2,…,c_i\}$ is the set of possible channel number of $i$-th layer, and $\times$ is the Cartesian product. The size of S is:

\begin{align}
    \|\mathcal{S}\| & = \|\mathcal{D}\|_1 \times \|\mathcal{D}\|_2 \times ... \times \|\mathcal{D}\|_L \notag\\
    & = \prod_{i=1}^{L}c_i \label{eq4}
\end{align}

In common CNNs such as ResNet and VGGNet, the channel number $c_i$ usually ranges from 16 to 1024, leading to $\prod_{i=1}^{L}c_i$  
being a large number. However, in practice we find that adding or reducing one channel in convolutional layers does 
not significantly affect the model accuracy. So it is natural to prune multiple channals at a time. We introduce a pruning 
step vector $E=[e_1,e_2,...,e_L],1 \leq e_i\leq c_i, e_i \in \mathbb{Z}$ to compress the search space. 
So the compressed search space $\mathcal{S}^{c}$ can be formulated as: 
\begin{equation}
    \mathcal{S}^{c} = \mathcal{D}^c_1 \times \mathcal{D}^c_2 \times ... \times \mathcal{D}^c_L \label{eq5}
\end{equation}

\noindent where $\mathcal{D}^c_i=\{e_i,2e_i,…,\lfloor \frac{c_i}{e_i} \rfloor e_i\}$, the size of $\mathcal{S}^{c}$ is:
\begin{align}
    \|\mathcal{S}^{c}\| & = \|\mathcal{D}^c\|_1 \times \|\mathcal{D}^c\|_2 \times ... \times \|\mathcal{D}^c\|_L \notag\\
    & = \prod_{i=1}^{L}\lfloor \frac{c_i}{e_i} \rfloor \label{eq6}
\end{align}

\begin{figure}[t]
    \centering
    \includegraphics[width=1\columnwidth]{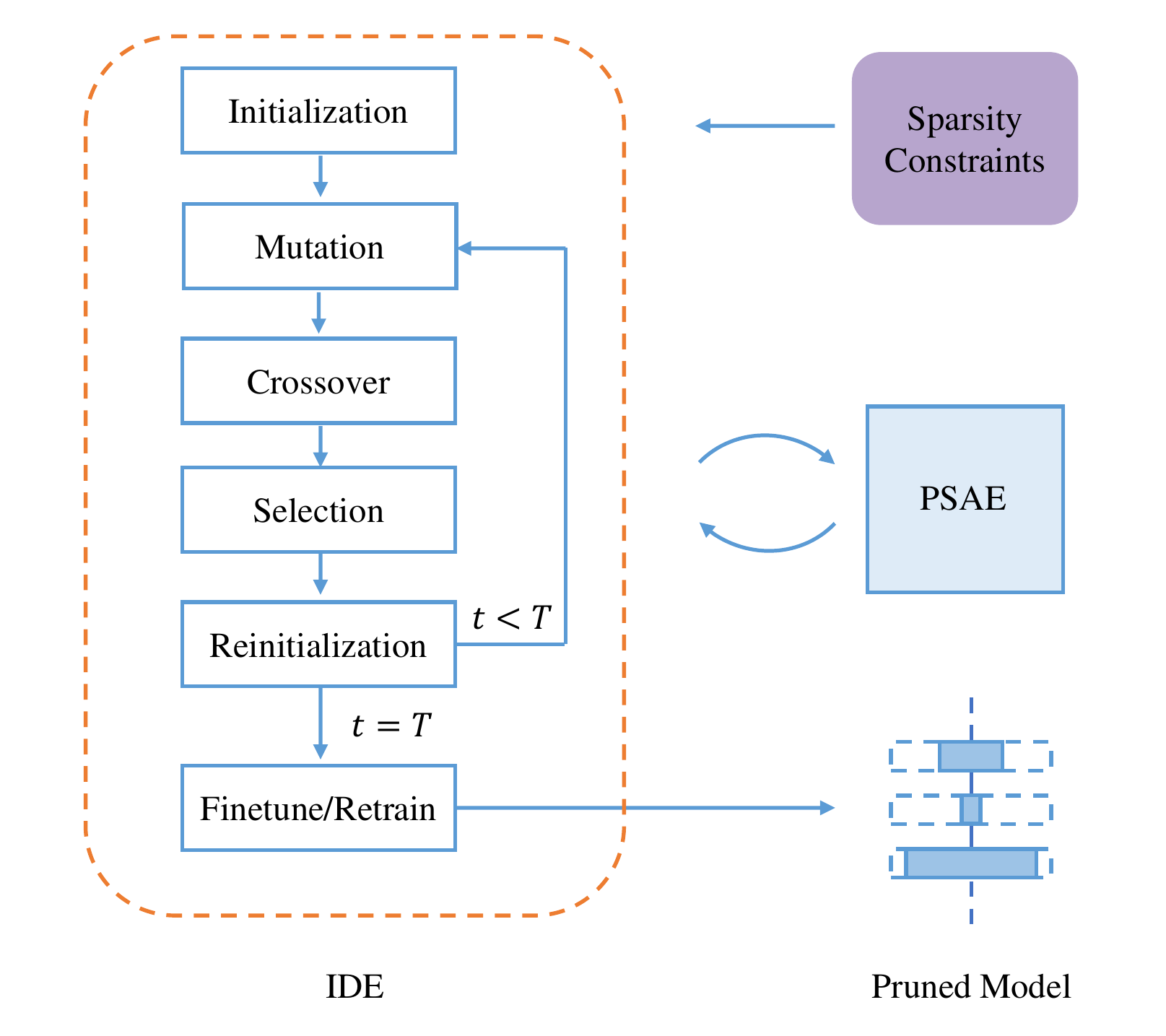} 
    \caption{
        The pipeline of IDE. Before iterations begin, we firstly initialize a population of structure vectors. Then the population will envolve in $T$ iterations. At each iteration, 
        we conduct Mutation, Crossover, Selection and Reinitialization in turn to every individual under sparsity constraints. After $T$ iterations, we output the optimal pruned structure and 
        fine-tune or retrain it to get the final pruned model.
    }
    \label{fig2}
\end{figure}

The benefits of introducing pruning step vector are: (1) Reducing the size of search space from $\prod_{i=1}^{L}c_i$ to 
$\prod_{i=1}^{L}\lfloor \frac{c_i}{e_i} \rfloor$. (2) The pruning step vector can be chosen freely according to the needs 
of different cases. For example, some tasks require that the channel number of a pruned model must be a multiple of 8, for hardware reasons. 
In this case, we can set $e_i = 8k, i=1,...,L, k \in \mathbb{N}$.

\subsection{The Improved Differential Evolution Algorithm}

Differential Evolution (DE) algorithm \cite{storn1997differential} is commonly used in evolutionary 
computing. We make some improvements based on the standard DE and then apply it to our problem. 
Our IDE algorithm is illustrated in Fig.\ref{fig2}. 
Before iterations begin, IDE will randomly initialize every individual. At each iteration, the population with $N$ individuals generates next generation by 
mutation, crossover, selection and reinitialization. After $T$ iterations, IDE outputs the optimal pruned architecture and fine-tunes or retrains the 
the optimal pruned structure to resume its accuracy.

\textbf{Population.} The population $X$ are consist of $N$ individuals $X_n,n=1,…,N$. Each individual $X_i$ is a 
structure vector. Hence searching for the individual of best fitness is equal to searching for the optimal structure vector. 
Considering an unpruned model $\mathcal{M}$ with structure vector $C=[c_1, c_2,..., c_L]$, the population of 
generation $t$ can be represented as:

\begin{align}
    X^t &= \{ X_1^t, X_2^t,..., X_N^t\} \notag\\
    X_n^t &\in \mathcal{S}^c, \quad n=1,2,...,N \label{eq7}
\end{align}

\noindent where $N$ is the size of the population, $X_n^t$ is the $n$-th individual of generation $t$ and 
$\mathcal{S}^c$ is the compressed search space.

\begin{figure*}[t]
    \centering
    \includegraphics[width=0.9\textwidth]{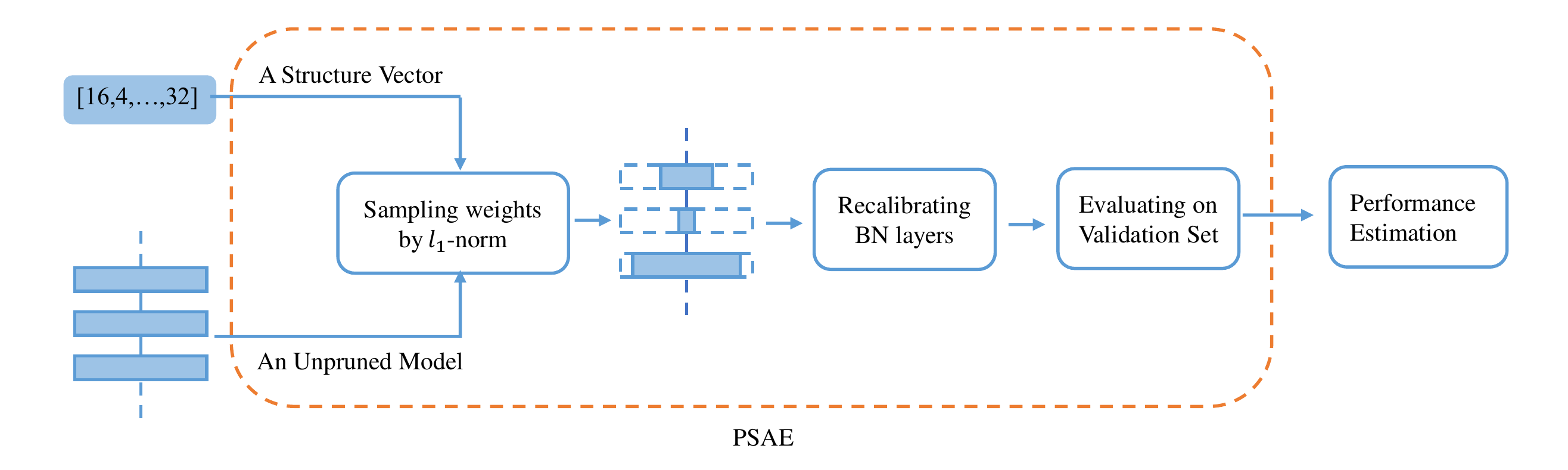} 
    \caption{
        The pipeline of PSAE. We firstly prepare an pretrained unpruned model $\mathcal{M}$ to sharing weights with pruned models. 
        Then, given a structure vector $C^{'}$, we will estimate its performance by: (1) Sharing weights from $\mathcal{M}$
        to $\mathcal{M^{'}}$, according to $l_1$-norm metric. (2) Recalibrating the statistics of BatchNorm layers of $\mathcal{M^{'}}$ with several thousands of samples
        from training dataset. (3) Output the accuracy of $\mathcal{M^{'}}$ on validation dataset as the estimation of $C^{'}$.
    }
    \label{fig4}
\end{figure*}

\begin{algorithm}[t]
	\renewcommand{\algorithmicrequire}{\textbf{Input:}}
	\renewcommand{\algorithmicensure}{\textbf{Output:}}
	\caption{Rescaling Structure Vector}
	\label{alg:1}
	\begin{algorithmic}[1]
        \REQUIRE Pruned structure vector $C^{'}=[c_{1}^{'},..., c_{L}^{'}]$, 
        unpruned structure vector $C=[c_{1},..., c_{L}]$,
        target of FLOP pruning rate $R_f$,
        target of parameter pruning rate $R_p$,
        pruning step vector $E=[e_1, ..., e_L]$.
		\ENSURE Rescaled structure vector $\hat{C}^{'}=[\hat{c}_{1},..., \hat{c}_{L}]$
        \FOR {each $i \in [1,L]$}
            \STATE $\hat{c}_{i}^{'} = \lfloor \frac{c_{i}^{'}}{e_{i}} \rfloor \times e_{i}$;
            \STATE if $\hat{c}_{i}^{'} < e_{i}$, $ \hat{c}_{i}^{'}=e_{i}$;
            \STATE if $\hat{c}_{i}^{'} > c_{i}$, $ \hat{c}_{i}^{'}=c_{i}$.
        \ENDFOR
        \WHILE{$r_f(\hat{C}^{'}) < R_f$ and $r_p(\hat{C}^{'}) < R_p$}
            \STATE Randomly select an index $ind$ from $[1,L]$;
            \STATE if $\hat{c}_{ind}^{'} > e_i$, $\hat{c}_{ind}^{'} = \hat{c}_{ind}^{'} - e_{i}$.
        \ENDWHILE
        \STATE \textbf{return} $\hat{C}^{'}=[\hat{c}_{1},..., \hat{c}_{L}]$.
	\end{algorithmic}  
\end{algorithm}

\textbf{Initialization.} The initial population $X^0$ is generated by randomly sampling in $\mathcal{S}^c$:
\begin{align}
    X^0 &= \{ X_1^0, X_2^0,..., X_N^0\} \notag\\
    X_n^0 &= \mathop{random} \limits_{C\sim p_{s}} (C), \quad n=1,2,...,N \label{eq8}
\end{align}

\noindent where $p_{s}$ is the distribution of compressed search space $\mathcal{S}^c$. However, an individual generated randomly may not satisfy the sparsity constraints.
In this case, we will rescale the individual until the sparsity constraints are satisfied, as illustrated in Alg.\ref{alg:1}: 
\begin{equation}
    \hat{X}_n^0 = rescale(X_n^0),\quad n=1,2,...,N \label{eq9}
\end{equation}

\noindent where $rescale(*)$ refers to Alg.\ref{alg:1}. After initialization, the population will evolve in $T$ iterations. At each iteration, we perform mutation, 
crossover, selection and reinitialization on all individuals.

\textbf{Mutation.} At iteration $t$, we randomly select three individuals $\hat{X}_p^t,\hat{X}_q^t$ and $\hat{X}_r^t$ from $X^t$. 
Then a new candidate of individual $\hat{V}_n^{t+1}$ can be generated by: 
\begin{align}
    V_n^{t+1} &= \hat{X}_p^t + F \times (\hat{X}_q^t-\hat{X}_r^t) \notag\\
    \hat{V}_n^{t+1} &= rescale(V_n^{t+1})  \label{eq10}
\end{align}

\noindent where $F\in[0,2]$ is the differential weight. In our experiment, $F$ is set to $0.5$.

\textbf{Crossover.} While mutation introduces individual-level diversity to the population, crossover introduces gene-level diversity. 
Crossover produces candidate individual by:
\begin{equation}
    U_{n,j}^{t+1} = \left\{  
        \begin{array}{rcl}  
        \hat{V}_{n,j}^{t+1},&   & if \quad rand < CR \\
        \hat{X}_{n,j}^t,&       & otherwise
    \end{array}  
\right.  \label{eq11}
\end{equation}

\noindent where $\hat{V}_n^{t+1}=[\hat{V}_{n,1}^{t+1},…,\hat{V}_{n,L}^{t+1}]$ is the candidate individual produced by mutation, 
$rand \in [0,1]$is a random number and $CR \in [0,1]$ is the crossover probability. In our experiments, $CR$ is set to $0.8$ and
$rand$ is generated randomly, subject to a uniform distribution.

\textbf{Selection.} In this step, we will compare the fitness of $U_{n}^{t+1}$ and $\hat{X}_{n}^t$, which is calculated by PSAE. 
If the fitness of $U_{n}^{t+1}$ is higher than $\hat{X}_{n}^t$, we update $X_{n}^{t+1}$ with $U_{n}^{t+1}$. Otherwise we keep 
$\hat{X}_{n}^t$ unchanged. We formulate selection as:
\begin{equation}
    X_{n}^{t+1} = \left\{  
        \begin{array}{rcl}  
        U_{n}^{t+1},& & if \quad PSAE(U_n^{t+1}) > PSAE(\hat{X}_n^t) \\
        \hat{X}_{n}^t,&  & otherwise
    \end{array}  
\right.  \label{eq12}
\end{equation}

\textbf{Reinitialization.} Since the initialization is conducted randomly, IDE may get stuck in locally optimal solutions. To make 
IDE more rubust to find the optimal solution, we propose to reinitialize those individuals who stay unchanged for 
$R$ generations/iterations. 
\begin{align}
    X_n^t &= \mathop{random} \limits_{C\sim p_{s}}(C), \quad if\quad X_n^t = X_n^{t-1} = ... = X_n^{t-R} \notag \\
    \hat{X}_n^t &= rescale(X_n^t) \label{eq13}
\end{align} 
\noindent where $p_{s}$ is the distribution of compressed search space $\mathcal{S}^c$. In our experiments, $R$ is set to 4. 

\begin{figure}[t]
    \centering
    \includegraphics[width=1\columnwidth]{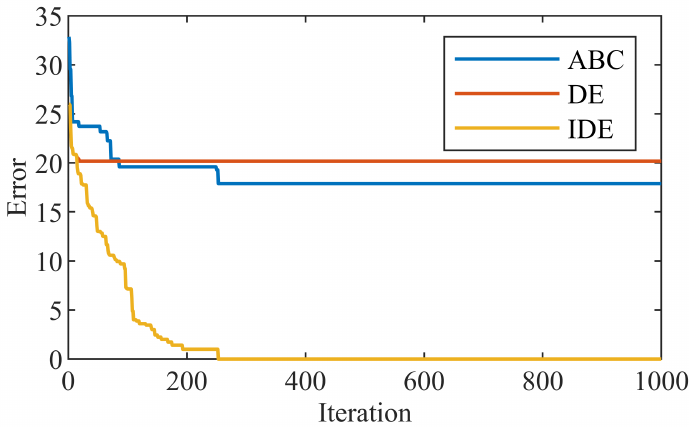} 
    \caption{
        Experimental results of Eq.\ref{eq14} by Differential Evolution algorithm (DE), Improved Differential Evolution algorithm (IDE) and Artificial
        Bee Colony algorithm (ABC \cite{lin2020channel}).
    }
    \label{fig3}
\end{figure}

\textbf{Disscussion.} Compared to DE, the improvement of IDE mainly lies in the Reinitialization step. Reinitialization
can speed up IDE and avoid IDE getting stuck in locally optimal solutions. To better illustrate advantages of Reinitialization, 
we utilize DE, IDE and ABC (Artificial Bee Colony algorithm \cite{lin2020channel}) to solve the following problem:
\begin{align}
    min \quad &||x-5||_2 \notag\\
    s.t. \quad &-10< x_i < 10, i=1,2,...,30 \notag\\
    &x_i \in \mathbb{Z}
    \label{eq14}
\end{align}
As illustrated in Fig.\ref{fig3}, IDE can find the optimal solution in less than 300 iterations, while DE and ABC cannot 
find optimal solution within 1000 iterations. This validates the effectiveness of IDE to search the optimal solution.

\subsection{Pruned Structure Accuracy Evaluator}

In evolution calculating, a set of candidates is optimized with regard to a given measure of fitness, a.k.a performance estimation strategy.
Those candidates of higher fitness are thought to be more important and will be kept. In this paper, we propose a novel performance estimation strategy,
named Pruned Structure Accuracy Evaluator (PSAE), to speed up the process of performance estimation.

As illustrated in Fig.\ref{fig4}, PSAE estimates the performance of a given structure vector by two steps: 
sampling weights from the unpruned model and recalibrating BatchNorm layers. Given a structure vector $C^{'}$, PSAE will initialize the corresponding model $\mathcal{M^{'}}$ by weights 
sampled from a pretrained unpruned model. The inherited weights are sorted by $l_1$-norm criterion instead of random selection, inspired by \cite{li2016pruning}. 
Since the statistics of BatchNorm layers of $\mathcal{M^{'}}$ are changed, we utilize several thousands of samples to recalibrate BatchNorm layers. 
Note that there is no training in this process and only around several thousands of samples from training dataset are used for recalibration. Thus calculating BatchNorm
post-statistics can be very fast. Finally, PSAE outputs the accuracy of $\mathcal{M^{'}}$ on validation dataset as the performance estimation of $C^{'}$. 
Our PSAE lies in the assumption that keeping "important" filters by $l_1$-norm metric can keep a large part of information in a model. 
More specifically, the relative ranking of accuracy among different structure vectors are what we are really interested in.

Compared to other performance estimation strategies, our PSAE has the advantage of faster performance estimation. We sampling weights from
the unpruned model instead of from an one-shot model. Training an unpruned model is simpler than training an one-shot model, for one-shot model has larger size and
more complex training pipeline. 

\begin{algorithm}[t]
	\renewcommand{\algorithmicrequire}{\textbf{Input:}}
	\renewcommand{\algorithmicensure}{\textbf{Output:}}
	\caption{Improved Differential Evolution Algorithm}
	\label{alg:2}
	\begin{algorithmic}[1]
        \REQUIRE Target of FLOP pruning rate $R_f$, target of parameter pruning rate $R_p$, differential weight $F$,
        crossover probability $CR$, iteration number $T$, population size $N$, pruning step vector $E$.
		\ENSURE The optimal pruned structure ${C}^{'*}$
        \STATE Initialize the Population $X^0 = \{X^0_{n}\}_{n=1,2,...,N}$ according to 
        Eq.\ref{eq8} and Eq.\ref{eq9};
        \FOR {each $t \in [1,T]$}
            \FOR {each $n \in [1,N]$}
                \STATE Perform Mutation according to Eq.\ref{eq10};\\
                \STATE Perform Crossover according to Eq.\ref{eq11};\\
                \STATE Perform Selection according to Eq.\ref{eq12}; \\
                \STATE Perform Reinitialization according to Eq.\ref{eq13}. \\
            \ENDFOR	
        \ENDFOR
		\STATE \textbf{return} ${C}^{'*} = \max \limits_{n=1,...,N} fitness(X^T_n)$.
	\end{algorithmic}  
\end{algorithm}

\section{Experiments}
\subsection{Implementation details}

\textbf{Datasets.} We conduct our experiments on both CIFAR10 \cite{krizhevsky2009learning} and ImageNet \cite{russakovsky2015imagenet} datasets.
To make it easier to compare with other methods, We study the performance of AACP on mainstream CNN models, including 
VGGNet, ResNet and MobileNetV2. Specifically, we prune VGG16/ResNet56/ResNet110 on CIFAR10 and ResNet50/MobileNetV2 on ImageNet.

\textbf{Training settings.} On CIFAR10, our training settings are the same as \cite{he2019filter}. We use SGD optimizer and step 
learning rate scheduler with momentum of $0.9$ and weight decay of 
$10^{-4}$. We train the unpruned model for $160$ epochs with an initial learning rate of $0.1$, and fine-tune the optimal pruned model 
for $160$ epochs with an initial learning rate of $0.01$. The learning rate is divided by $10$ at $80$-th epoch and $120$-th epoch.  

On ImageNet, our data argumentation strategies are the same as PyTorch \cite{paszke2017automatic} official examples. 
For ResNet50 on ImageNet, SGD optimizer and step learning rate scheduler are used. 
We train the unpruned model for $90$ epochs with an initial learning rate of $0.1$ and fine-tune the 
optimal pruned model for $90$ epochs with an initial learning rate of $0.01$. The learning rate 
is divided by $10$ every $30$ epochs. 
For MobileNetV2 on ImageNet, we follow the settings in \cite{wang2019pruning}.
We use cosine learning rate scheduler with an initial learning rate of $0.05$, momentum of $0.9$, weight decay of $0.00004$.    
SGD optimizer is used and the total training epochs is $150$.
On CIFAR10, models are trained and fine-tuned on one GTX 2080 GPU with a batch size of $64$. 
On ImageNet, 8 GTX 2080 GPUs are used. The batch size for ResNet50 is $512$ and batch size for MobileNetV2 is 256.

\textbf{Pruning settings.} The first step of AACP is to set the pruning step vector 
$E=[e_1,e_2,...,e_L],1 \leq e_i \leq c_i$. In our experiments, the pruned step $e_i$ is set to one eighth 
of channel number $c_i$. Take VGG16 as an example, the structure vector $C=[64,64,128,128,256,256,256,512,512,512,
512,512,512]$, so $E=[8,8,16,16,32,32,32,64,64,64,64,64,64]$. Note that the value of $e_i$ is flexible according 
to user selection. After compressing search space, AACP utilizes IDE to search an optimal pruned structure. 
Here we set the population size $N=10$, differential weight $F=0.5$ and crossover probability $CR=0.8$ empirically.

\subsection{Visualization of Pruning}

\begin{figure}[t]
    \centering
    \includegraphics[width=1\columnwidth]{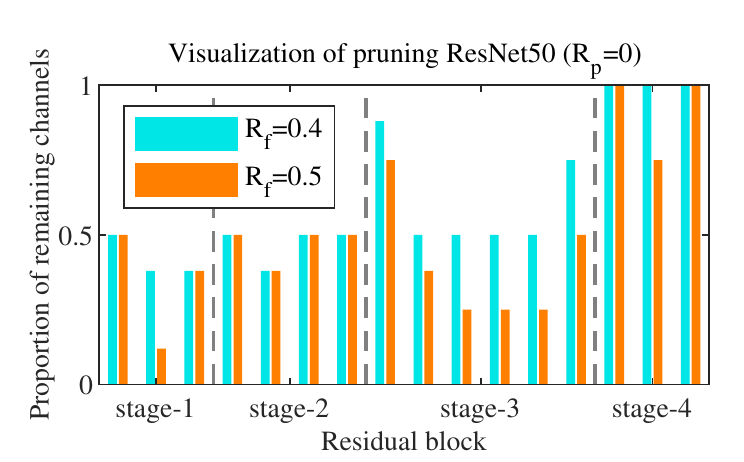} 
    \caption{
        Visualization of pruning ResNet50 on ImageNet with $R_p = 0$ and $R_f=0.4/0.5$. ResNet50 has four stages. Different stages have different sizes of 
        feature maps. Each stage has $3$ to $6$ residual blocks. We draw the proportion of remaining channels of each block. An unpruned model is $100\%$ reserved for every block. 
        Best viewed in color.
    }
    \label{fig5}
\end{figure}

In this part, we visualize our optimal structure vectors and discuss some insights based on the results.
We prune ResNet50 on ImageNet with a fixed $R_p = 0$ (no constraints on parameters)
, while changing $R_f$ from $0.4$ to $0.5$. The pruned results are illustrated
in Fig.\ref{fig5}. When $R_f$ is set to $0.4$ (showed in blue), layers in stage-1 and stage-2 keep no more than $50\%$
channels, while layers in stage-4 keep more than $70\%$ channels. This imples when not restricting parameter pruning rate, 
our method tends to prune shallower blocks to reduce FLOPs and keeps more channels in deeper blocks.
When $R_f$ is set to $0.5$ (showed in orange), we observe the same phenomenon of keeping more channels in deeper blocks. We also find that 
when increasing $R_f$ from $0.4$ to $0.5$, our method chooses to prune more channels of those
blocks which are in the middle of a stage, while keep more channels of first block in each stage. This results in a U-shaped curve within a stage, 
as illustrated in Fig.\ref{fig5}. We think this is because the first block of every stage downsamples the feature map and thus requires 
more channels to avoid information loss.

\begin{figure}[t]
    \centering
    \includegraphics[width=0.98\columnwidth]{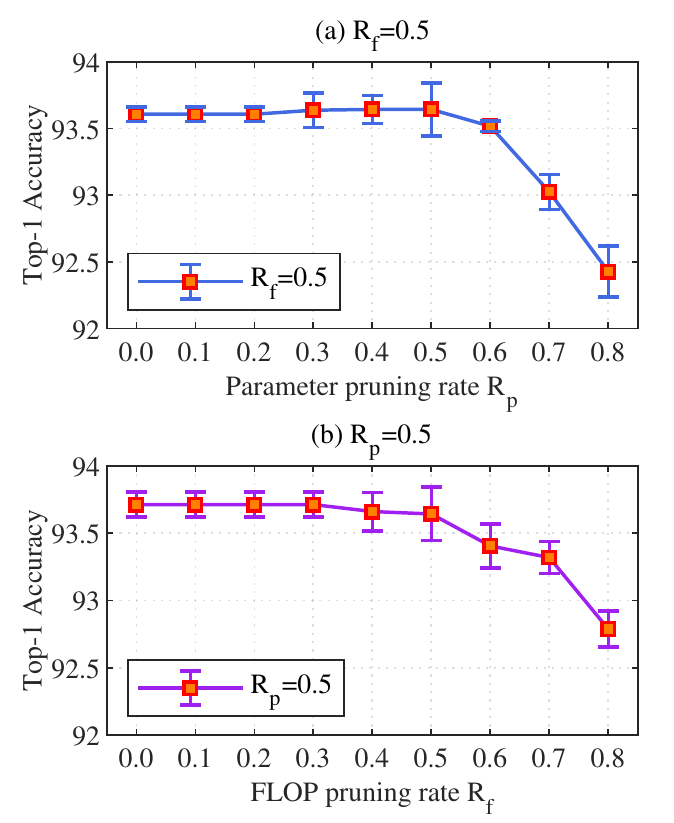} 
    \caption{
        Visualization of pruning VGG16 on CIFAR10 dataset. In (a), we fix $R_f=0.5$ and change $R_p$ from $0$ to $0.8$. 
        In (b), we fix $R_p=0.5$ and change $R_f$ from $0$ to $0.8$.
    }
    \label{fig6}
\end{figure}

We also explore the relationship between $R_f$ and $R_p$ by pruning VGG16 on CIFAR10 dataset. Firstly, we fix $R_f=0.5$
and change $R_p$ from $0$ to $0.8$. Fig.\ref{fig6}(a) shows the top-1 accuracy of optimal pruned structures.
When $R_f=0.5, 0 < R_p < 0.5$, the top-1 accuracy of optimal pruned structures doesn’t decrease, which even increase slightly. This implies
that $R_f$ is the stricter constraint compared to $R_p$ when $R_f=0.5, 0 < R_p < 0.5$. However, When $R_f=0.5, R_p > 0.5$, the 
curve goes down, which means $R_p$ is the stricter constraint in this case. Secondly, we fix $R_p=0.5$ and change $R_f$ from 
$0$ to $0.8$. Fig.\ref{fig6}(b) shows the same trend as Fig.\ref{fig6}(a). These curves reveals that the optimal structures are controlled 
by both $R_f$ and $R_p$. By changing $R_f$ and $R_p$, our AACP can accurately compress a model from multiple levels.

\subsection{Comparisons with state-of-the-art}

In this section, we compare our AACP with other channel pruning methods, including uniform channel number shrinkage (uniform), 
$l_1$-norm pruning method (L1-norm) \cite{li2016pruning}, Network Slimming (NS) \cite{liu2017learning},
ThiNet \cite{luo2017thinet}, Channel Pruning (CP) \cite{he2017channel}, Discrimination-aware Channel Pruning (DCP) \cite{zhuang2018discrimination},
Pruning From Scratch (PFS) \cite{wang2019pruning}, ABCPruner \cite{lin2020channel}, reinforcement learning method (AMC) \cite{he2018amc}, 
SFP \cite{he2018soft}, Rethinking \cite{liu2018rethinking}, FPGM \cite{he2019filter}, DMCP \cite{guo2020dmcp}, AutoSlim \cite{yu2019autoslim} and 
MetaPruning \cite{liu2019metapruning}. To eliminate the influences of different training settings and experiment environments, we mainly compare the top-1 accuracy drop rate with other methods 
under the same FLOP pruning rate. To make it fairer, we only impose FLOP pruning rate $R_f$ and set $R_p=0$ in this section, like other channel pruning methods. 

\begin{table}[tb]
    \centering
    \caption{Results on CIFAR10 dataset. S represents training $C^{'*}$ from scratch and F
    represents fine-tuning $C^{'*}$ by inheriting weights. $R_p = 0$ for all results. 
    $R_f$ is anotated in the table, e.g. (F,$0.50$) means fine-tuning $C^{'*}$ and $R_f = 0.50$.
    }
    \resizebox{\columnwidth}{!}{
      \begin{tabular}{c|ccccc}
      \toprule
      CNN   & Method & $\Delta$FLOPs & Baseline(\%) & Pruned(\%) & $\Delta$Acc(\%) \\
      \midrule
      \multirow{11}[2]{*}{\rotatebox{90}{VGG16}} 
            & L1-norm   & -34\% & 93.25 & 93.40 & +0.15 \\
            & NS   & -51\% & 93.99 & 93.80 & -0.19 \\
            & ThiNet & -50\% & 93.99 & 93.85 & -0.14 \\
            & CP    & -50\% & 93.99 & 93.67 & -0.32 \\
            & DCP   & -50\% & 93.99 & 94.16 & +0.17 \\
            & PFS   & -50\% &93.44 & 93.63 $\pm$0.06 & \textbf{+0.19} \\
            & ABCPruner & -73\% & 93.02 & 93.08 & \textbf{+0.06} \\
            & Ours (S,$0.50$) & -50\% & 93.56$\pm$0.19 & 93.72$\pm$0.13 & +0.16 \\
            & Ours (F,$0.50$) & -50\% & 93.56$\pm$0.19 & 93.61$\pm$0.05 & +0.05 \\
            & Ours (S,$0.70$) & -70\% & 93.56$\pm$0.19 & 93.57$\pm$0.12 & +0.01 \\
            & Ours (F,$0.70$) & -70\% & 93.56$\pm$0.19 & 93.39$\pm$0.07 & -0.17 \\
      \midrule
      \multirow{11}[2]{*}{\rotatebox{90}{ResNet56}} 
            & Uniform & -50\% & 92.80 & 89.80 & -3.00 \\
            & ThiNet & -50\% & 93.80 & 92.98 & -0.82 \\
            & CP    & -50\% & 93.80 & 92.80 & -1.00 \\
            & DCP  & -50\% & 93.80 & 93.49 & -0.31 \\
            & AMC   & -50\% & 92.80 & 91.90 & -0.90 \\
            & SFP   & -50\% & 93.59 & 93.35$\pm$0.31 & -0.24 \\
            & Rethink & -50\% & 93.80 &93.07$\pm$0.25 & -0.73 \\
            & PFS & -50\% & 93.23 & 93.05$\pm$0.19 & -0.18 \\
            & ABCPruner & -54\% & 93.26  & 93.23 & -0.03 \\
            & Ours (S,$0.50$) & -50\% & 93.10$\pm$0.20&93.31$\pm$0.28& \textbf{+0.21} \\
            & Ours (F,$0.50$) & -50\% & 93.10$\pm$0.20&92.82$\pm$0.06& -0.28 \\
      \midrule
      \multirow{9}[2]{*}{\rotatebox{90}{ResNet110}} 
            & L1-norm    & -40\%  & 93.53 & 93.30 & -0.23 \\
            & SFP   & -40\% & 93.68 & 93.86$\pm$0.30 & +0.18 \\
            & Rethink & -40\% & 93.77 & 93.92$\pm$0.13 & +0.15 \\
            & PFS   & -40\% & 93.49 & 93.69$\pm$0.28 &+0.20 \\
            & ABCPruner & -65\% & 93.50 & 93.58 & +0.08 \\
            & Ours (S,$0.40$) &-40\% & 93.30$\pm$0.08 & 93.72$\pm$0.43 & +0.42 \\
            & Ours (F,$0.40$) &-40\% & 93.30$\pm$0.08 & 93.76$\pm$0.16 & \textbf{+0.46} \\
            & Ours (S,$0.65$) &-65\% & 93.30$\pm$0.08 & 93.56$\pm$0.12 & \textbf{+0.26} \\
            & Ours (F,$0.65$) &-65\% & 93.30$\pm$0.08 & 93.33$\pm$0.10 & +0.03 \\
      \bottomrule
      \end{tabular}%
    }
    \label{table1}%
\end{table}%

\begin{table}[htb]
    \centering
    \caption{Pruning results on ImageNet dataset. S represents training $C^{'*}$ from scratch and F
    represents fine-tuning $C^{'*}$ by inheriting weights. $R_p = 0$ for all results. 
    $R_f$ is anotated in the table, e.g. (F,$0.42$) means fine-tuning $C^{'*}$ and $R_f = 0.42$.
    }
    \resizebox{\columnwidth}{!}{
      \begin{tabular}{c|ccccc}
      \toprule
      CNN   & Method & $\Delta$FLOPs & Baseline(\%) & Pruned(\%) & $\Delta$Acc(\%) \\
      \midrule
      \multirow{10}[2]{*}{\rotatebox{90}{ResNet50}} 
            & ThiNet-70 & -36.8\% & 72.88 & 72.04 & -0.84 \\
            & FPGM  & -42.2\% & 76.15 & 75.59 & -0.56 \\
            & DMCP  & -46.3\% & 76.60 & 76.20 & -0.40 \\
            & PFS   & -51.2\% & 77.20 & 75.60 & -1.60 \\
            & FPGM  & -53.5\% & 76.15 & 74.83 & -1.32 \\
            & AutoSlim & -51.2\% & 76.10 & 75.60 & -0.50 \\
            & ThiNet-50 & -55.8\% & 72.88 & 71.01 & -1.87 \\
            & MetaPruning & -51.2\% & 76.6 & 75.4 & -1.20 \\ 
            & ABCPruner & -54.3\% & 76.01 & 73.86 & -2.15 \\
            & Ours (S,$0.42$) & -42.0\% & 75.94 & 75.19 & -0.75 \\
            & Ours (F,$0.42$) & -42.0\% & 75.94 & 75.77 & \textbf{-0.18} \\
            & Ours (S,$0.51$) & -51.7\% & 75.94 & 75.01 & -0.93\\
            & Ours (F,$0.51$) & -51.7\% & 75.94 & 75.46 & \textbf{-0.48} \\
      \midrule
      \multirow{7}[2]{*}{\rotatebox{90}{MobileNetV2}} 
            & Uniform 1.0x & -0.0\% & 71.8 & 71.8 & -0.0 \\
            & Uniform 0.75x & -30.0\% & 71.8 & 69.3 & -2.5  \\
            & PFS & -30.0\% & 72.1 & 70.9 & -1.2 \\
            & MetaPruning & -27.7\% & 72.0 & 71.2 & -0.8 \\
            & AMC  & -29.7\% & 71.8 & 70.8 & -1.0 \\
            & AutoSlim  & -29.7\% & 74.2 & 73.0 & -1.2 \\
            & DMCP   & -29.7\% & 74.6 & 73.5 & -1.1 \\
            & Ours (S,$0.30$) & -30.2\% & 71.8 & 70.7 & -1.1 \\
            & Ours (F,$0.30$) & -30.2\% & 71.8 & 71.1 & \textbf{-0.7} \\
      \bottomrule
      \end{tabular}%
    }
    \label{table2}%
\end{table}%

We prune VGG16/ResNet56/ResNet110 on CIFAR10 and prune ResNet50/MobileNetV2 on ImageNet.
After searching the optimal pruned structure $C^{'*}$, we utilize two ways to resume the accuracy of 
$C^{'*}$: (1) Fine-tuning $C^{'*}$ by 160 epochs on CIFAR10 and 90 epochs on ImageNet (denoted as F); (2) Training $C^{'*}$
from scratch by 320 epochs on CIFAR10 and 180 epochs on ImageNet (denoted as S). 

The results on CIFAR10 dataset are illustrated in Table.\ref{table1}. We run each experiment five times and report the “mean
± std”. When pruning VGG16, our method reduces $50\%$ FLOPs 
of VGG16 with a raise of $0.16\%$ top-1 accuracy, which is slightly worse than PFS \cite{wang2019pruning} 
but better than L1-norm \cite{li2016pruning}, NS \cite{liu2017learning}, ThiNet \cite{luo2017thinet}, CP \cite{he2017channel} 
and DCP \cite{zhuang2018discrimination}. We also achieve comparable results to ABCPruner \cite{lin2020channel} when reducing $70\%$ FLOPs of VGG16. 
When pruning ResNet56 on CIFAR10, our method reduces $50\%$ FLOPs of ResNet56 with 
a raise of $0.21\%$ top-1 accuracy, which surpasses other methods. As for pruning ResNet110 on CIFAR10, 
we outperform other methods when reducing $40\%$ FLOPs and $65\%$ FLOPs.

We also compare AACP to other channel pruning methods on ImageNet. Table.\ref{table2} illustrates the results of pruning ResNet50
and MobileNetV2 on ImageNet. When reducing $42.0\%$ FLOPs of ResNet50, our method only causes $0.18\%$ top-1 accuracy drop, 
which is better than ThiNet-70 \cite{luo2017thinet}, FPGM \cite{he2019filter}, DMCP \cite{guo2020dmcp}. When reducing $51.7\%$ FLOPs of ResNet50, 
our method only causes $0.48\%$ top-1 accuracy drop, 
which is better than PFS \cite{wang2019pruning}, FPGM \cite{he2019filter}, AutoSlim \cite{yu2019autoslim}, ThiNet-50 \cite{luo2017thinet}, 
MetaPruning \cite{liu2019metapruning} and ABCPruner \cite{lin2020channel}. For MobileNetV2, we reduce $30\%$ FLOPs with $0.7\%$ drop 
of top-1 accuracy, which surpasses other methods. In most cases of our experiments, our AACP outperforms other channel pruning methods.
This validates that our method discovers more powerful pruned structures in those cases.

\subsection{Ablation study}

\textbf{Influence of $l_1$-norm metric}. To study the effectiveness of $l_1$-norm metric used in PSAE, 
we use random selection to replace $l_1$-norm and compare their results. 
We prune VGG16 and ResNet56 on CIFAR10, with $l_1$-norm and random selection respectively.
As Table.\ref{table3} illustrates, $l_1$-norm metric outperforms random selection. This validates that $l_1$-norm is an 
effective criterion to select "important" filters that contain more information of a model when estimating the performance
of a given architecture.

\begin{table}[tbp]
    \centering
    \caption{Results of pruning VGG16/ResNet56 on CIFAR10 by random selection and $l_1$-norm.}
    \resizebox{\columnwidth}{!}{
      \begin{tabular}{clrrrr}
      \toprule
      \multicolumn{1}{l}{Model} & metric & \multicolumn{1}{c}{$\Delta$FLOPs} & \multicolumn{1}{c}{Baseline} & \multicolumn{1}{c}{Pruned} &\multicolumn{1}{c}{$\Delta$Acc} \\
      \midrule
      \multirow{2}[2]{*}{VGG16} 
            & random & -50.4\% & 93.56$\pm$0.19 & 93.34$\pm$0.15& -0.22 \\
            & $l_1$-norm & -50.4\%  & 93.56$\pm$0.19 & 93.61$\pm$0.05 & +0.05 \\
      \midrule
      \multirow{2}[2]{*}{ResNet56} 
            & random & -50.1\%  & 93.10$\pm$0.20& 93.21$\pm$ 0.09 & +0.11\\
            & $l_1$-norm &  -50.1\%  & 93.10$\pm$0.20 & 93.31 $\pm$ 0.28& +0.21 \\
      \bottomrule
      \end{tabular}%
    }
    \label{table3}%
  \end{table}%

\begin{table}[t]
    \centering
    \caption{The results of pruning three pre-trained model: $\mathcal{M}_\alpha$,$\mathcal{M}_\beta$ and $\mathcal{M}_\gamma$
            when $R_f=0.5$, $R_p=0$. "F" refers to fine-tuing the optimal pruned structure. "S" refers to
            training the optimal pruned structure from scratch.
            }
    \resizebox{\columnwidth}{!}{
      \begin{tabular}{ccccc}
      \toprule
        Model & Epoch & Unpruned Acc & Pruned Acc (F) & Pruned Acc (S)  \\
      \midrule
      $\mathcal{M}_\alpha$     & 30    & 57.57\% & 75.00\% & 74.90\%\\
      $\mathcal{M}_\beta$      & 60    & 72.26\% & 75.20\% & 74.92\%\\
      $\mathcal{M}_\gamma$     & 90    & 75.94\% & 75.46\% & 75.01\%\\
      \bottomrule
      \end{tabular}%
    }
    \label{table4}%
  \end{table}%

\textbf{Influence of a well-trained unpruned model}. To study the effect of a well-trained unpruned model, 
we apply our AACP on three unpruned models, which are $\mathcal{M}_\alpha$(trained for 30 epochs), 
$\mathcal{M}_\beta$(trained for 60 epochs) and $\mathcal{M}_\gamma$ (trained for 90 epochs). 

The optimal pruned structures of $\mathcal{M}_\alpha, \mathcal{M}_\beta$ and $\mathcal{M}_\gamma$ are
$C^{'*}_\alpha, C^{'*}_\beta, C^{'*}_\gamma$ respectively. As Table.\ref{table4} illustrates, 
the accuracy of training $C^{'*}_\alpha, C^{'*}_\beta, C^{'*}_\gamma$ from scratch (Table.\ref{table4}, Pruned Acc (S)) is very close.
This implies that to get an optimal pruned structure, we don't have to train an unpruned model fully. However, when we fine-tune
$C^{'*}_\alpha, C^{'*}_\beta$ and $C^{'*}_\gamma$ instead of training them from scratch, the rank of their accuracy is 
$C^{'*}_\alpha < C^{'*}_\beta < C^{'*}_\gamma$ (Table.\ref{table4}, Pruned Acc (F)). This means an well-trained unpruned model ($\mathcal{M}_\gamma$)
is helpful for offering better initialized
weights for pruned model, because the structure of $C^{'*}_\gamma$ is no better than the others.
This finding is not contradictory with the conclusion of Rethinking \cite{liu2018rethinking}, for our comparison is implemented by fine-tuning a pruned model 
in fixed epochs from different unpruned models. While in Rethinking \cite{liu2018rethinking}, the epochs of fine-tuning a model are less than training it from scratch.

\section{Conclusion}

In this paper, we propose a novel channel pruning method AACP to prune CNNs in an end-to-end manner, which achieves state-of-the-art 
performance in many pruning cases. Our method largely speeds up the process of performance estimation and can deal with multiple 
sparsity constraints to realize accuracy and automatic prunning. In the future work, we will explore how to extend the search space and apply AACP to more tasks.

{\small
\bibliographystyle{ieee_fullname}
\bibliography{mybib}
}

\end{document}